# Local-peak scale-invariant feature transform for fast and random image stitching

Hao Li[1], Lipo Wang[2], Tianyun Zhao[3], Wei Zhao[1,*]

1. State Key Laboratory of Photon-Technology in Western China Energy, International Collaborative Center on Photoelectric Technology and Nano Functional Materials, Institute of Photonics & Photon Technology, Northwest University, Xi'an 710127, China
2. UM-SJTU Joint Institute, Shanghai Jiao Tong University, Shanghai 200030, China
3. School of Automation, Northwestern Polytechnical University, Xi'an 710072, China

Correspondence: zwbayern@nwu.edu.cn

**Abstract**

Image stitching aims to construct a wide field of view with high spatial resolution, which cannot be achieved in a single exposure. Typically, conventional image stitching techniques, other than deep learning, require complex computation and thus computational pricy, especially for stitching large raw images. In this study, inspired by the multiscale feature of fluid turbulence, we developed a fast feature point detection algorithm named local-peak scale-invariant feature transform (LP-SIFT), based on the multiscale local peaks and scale-invariant feature transform method. By combining LP-SIFT and RANSAC in image stitching, the stitching speed can be improved by orders, compared with the original SIFT method. Nine large images (over 2600×1600 pixels), arranged randomly without prior knowledge, can be stitched within 158.94 s. The algorithm is highly practical for applications requiring a wide field of view in diverse application scenes, e.g., terrain mapping, biological analysis, and even criminal investigation.

**1. Introduction**

Image stitching is employed to reconstruct complete image information from its fragments [1]. Due to the limited capability to capture a large area with high spatial resolution, image stitching finds widespread uses in engineering applications such as machine vision [2-5], augmented reality [6-9], navigation [10, 11] and panoramic shooting [12-14] etc. Furthermore, it plays an indispensable role in supporting scientific research, such as the construction of large-scale bioimages [15-19], micro/nanostructures [20-22], aerial photography [23-25] and space remote sensing [26] etc.

Various image stitching techniques can be traced back to the late 1970s or early 1980s [27-29] aiming to merge multiple images to achieve a wider field of view. In that early stages, multiple images of the same scene were simply superimposed [30]. To now, advanced image stitching techniques can be primarily categorized into two categories: region-based and feature-based [28]. In comparison to the computationally intensive region-based stitching techniques [13], feature-based techniques impose fewer requirements on region overlapping [31], rendering them increasingly appealing in recent years. Harris et al. [32] initially proposed a feature-based stitching technique for detecting features in images based on corner points. Subsequently, Lowe et al. [33] introduced the scale-invariant Feature Transform (SIFT) approach for feature detection, utilizing the Gaussian pyramid and Gaussian difference pyramid. Although SIFT has a series of features, such as robustness, scale invariance and rotation invariance, it necessitates substantial computational resources, leading to a slower feature point detection, and accordingly, lower stitching speed.

To address the limitations of SIFT, Rosten et al. [34] proposed the Features from Accelerated Segment Test (FAST), a machine learning algorithm for high-speed corner detection. However, FAST sacrifices scale invariance and rotation invariance features. To mitigate these drawbacks, Bay et al. [35] proposed the Speeded Up Robust Features (SURF) algorithm in 2008, which introduced a new scale and rotation invariant detector and descriptor. While SURF offers scale invariance, and rotation invariance, it is less robust to viewpoint changes. In 2011, Rublee et al. [36] presented an efficient alternative to SIFT and SURF, named as Oriented FAST and Rotated BRIEF (ORB). However, it is susceptible to variations in illumination. In the same year, Stefan et al. [37] proposed Binary Robust Invariant Scalable Keypoints (BRISK) method. Similar to SIFT, BRISK predominantly utilizes FAST9-16 to identify feature points and generates feature descriptor vectors in binary form, which exhibit sensitivity to variations in illumination and rotation. Later, to overcome boundary blurring and detail loss in SIFT and SURF, Alcantarilla et al. [38] proposed KAZE algorithm in 2012, which can preserve more details by constructing a nonlinear scale space to detect feature points. However, similar to SIFT, KAZE algorithm also requests a large calculation time. In 2013, Pablo et al. [39] proposed Accelerated KAZE (AKAZE) algorithm to improve the calculation efficiency by changing the description of feature vectors.

Although in recent years, the research focuses are shifting more toward the deep-learning neural network [40-42], it is still effective to optimize and update the conventional algorithms for image stitching applications. For instance, the parity algorithm by Wu et al. [43], as an improvement of SIFT, can reduce the matching error with better accuracy. The segmentation algorithm by Gan et al. [44] helps to improve the stitching speed and the SIFT accuracy. An image registration method [45] combines Shi-Tomasi with SIFT to improve the matching accuracy and reduce the computational cost.

In feature-based image stitching techniques, the precise definition and detection of feature points are pivotal. The algorithm's efficiency can be significantly enhanced if there is a reasonable replacement for the Gaussian pyramid and Gaussian difference pyramid operations in SIFT. In the Dissipation Element (DE) analyses of turbulent flows [46, 47], the dynamic nature of the field can be characterized by the statistics of the statistical properties of the extremal points. Moreover, the concept of extremal points can be extended to the multiscale level [48]. From a geometric standpoint, any image can be viewed as a structure with multiscale intensity, akin to the identified DEs. Considering the multiscale nature of the DE analysis in turbulent flow field and that in image analysis, it is possible to replace the Gaussian pyramid in the SIFT algorithm by the extremal points in DE analysis, to improve the efficiency of feature point detection in SIFT algorithm.

In this study, we aim to construct a fast feature point detection algorithm, termed Local-Peak Scale-Invariant Feature Transform (LP-SIFT), which integrates SIFT with the concept of local extremal points or image peaks at the multiscale level. Specifically, the featured points can be substituted with the multiscale local peak points without reliance on the Gaussian pyramid and differential pyramid in SIFT. Consequently, the efficiency of feature point detection can be significantly enhanced. Additionally, by combing LP-SIFT and RANSAC in image stitching, the stitching speed can be notably improved, particularly in the processing of large images, when compared with the original SIFT method. The framework of the algorithm is briefly introduced in Fig. 1. Finally, we demonstrate that random images or segments of a large image can be successfully stitched within an acceptable time without requirement of prior knowledge.

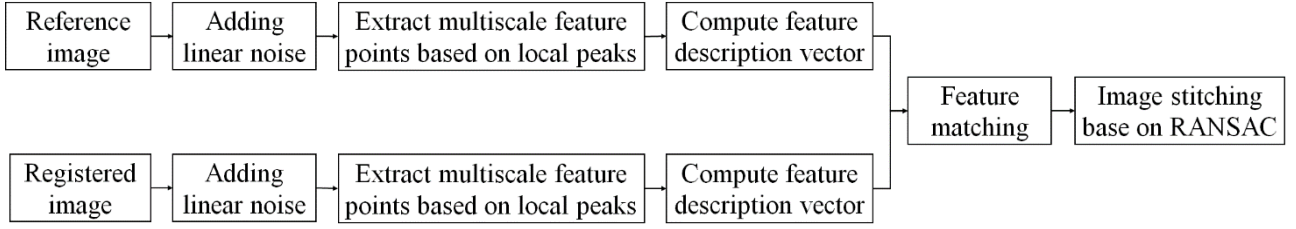

Figure 1. Framework of the LP-SIFT method.

## 2. Materials and methods

The principle of the LP-SIFT method is schematically diagramed in Fig. 2. Generally, there are five sub steps, including image preprocessing, feature point detection, feature point description, feature point matching and image stitching. We will initially provide a detailed introduction to each step. Subsequently, a strategy of applying LP-SIFT to restore a large image from random image fragments will be elucidated.

2.1 Image preprocessing

In the following exemplified case, the two images (Fig. 2(a)) designated for stitching are referred to as the reference image (stored as the reference matrix $M_1$) and registered image (stored as the registration matrix $M_2$), respectively. Since the feature points adopted in LP-SIFT are the local peak points (both maximum and minimum), to avoid the difficulty to locate them in the constant image intensity region, e.g. due to saturation, it is numerically meaningful to first impose a small linear background to both images. Therefore, the new image matrixes $M_{n,k}$ becomes

$$M_{n,k}(i,j) = M_k(i,j) + [(i-1)*nc_k + j]*\alpha \tag{1}$$

where $\alpha \ll 1$ is the linear noise coefficient. Now the size of $M_{n,k}$ is $nr_k \times nc_k$, $nr_k$ is the row of the $M_{n,k}$, $nc_k$ is the column of the $M_{n,k}$, where $k=1$ represents parameters of the reference image and $k=2$ represents parameters of the registered image.

2.2 Feature point detection

In step 2, we utilize the local peak points of multiscale images as the feature points (Fig. 2(b)). Both the reference image and the registered image are partitioned into squared interrogation windows of size $L$, with varying scales, e.g. $L=32$ to $128$. The maximum and minimum points in each interrogation window are collected as feature points, which can be formulated as

$$\begin{cases} M_{n,k,max}(p,L) = max[M_{np,k}(i,j)], i,j \in [0,L] \\ M_{n,k,min}(p,L) = min[M_{np,k}(i,j)], i,j \in [0,L] \end{cases} \tag{2}$$

where $M_{np,k}(i,j)$ is the $p$th interrogation window of $L \times L$ pixels.

2.3 Feature point description

Once all the feature points are gathered, the SIFT feature description vector is employed to characterize the acquired feature points. First, as shown in Fig. 2(d), around each feature point (represented by red spot), a square region with the size $w$ is extracted, with

$$w = \beta L d, \tag{3}$$

where $\beta = \beta_0 (L/L_{max})^{-1/2}$ is an adjustable factor to control the size of the square region and $\beta_0$ is the initial $\beta$ when the interrogation window size $L$ equals to its maximum value $L_{max}$ (e.g. 128). To ensure $w$ is not over $L$, it requires

$$\beta_0 \leq \frac{1}{d}\left(\frac{L}{L_{max}}\right)^{-\frac{1}{2}}. \tag{4}$$

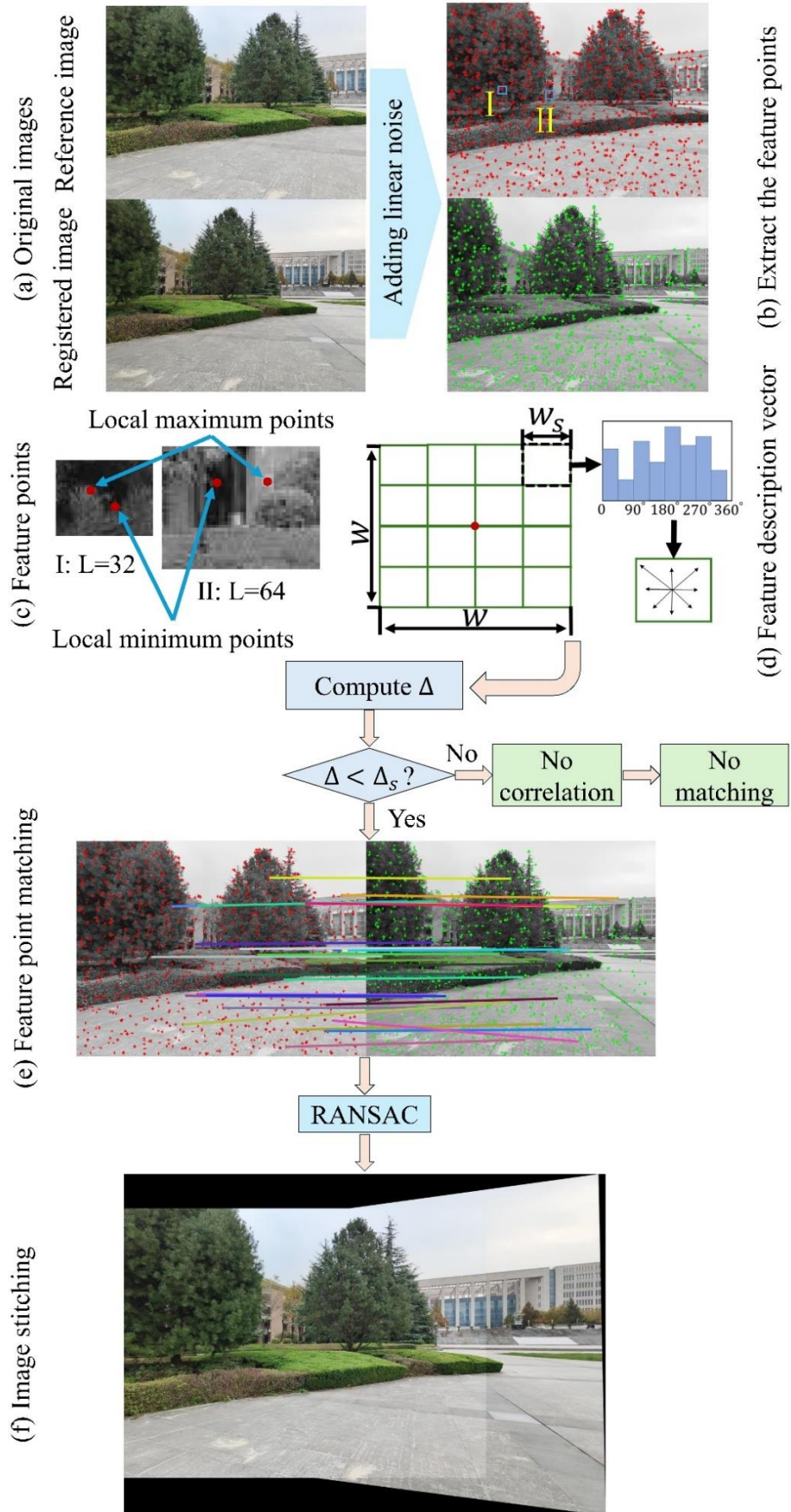

Figure 2. Diagram of the LP-SIFT method.

We further divide the square region into $d \times d$ subregions, e.g. $d = 4$ as commonly used [49]. In each subregion $M_{n,k,s}$, there are $w_s \times w_s$ pixels, with $w_s = \beta L$ as the side length of the subregion. At each pixel $(i,j)$ in this subregion, the magnitude $f_k(i,j)$ and orientation $\theta_k(i,j)$ of the image gradient can be calculated as

$$\begin{cases} f_k(i,j) = \sqrt{a^2 + b^2} \\ \theta_k(i,j) = b/a \\ a = M_{n,k,s}(i+1,j) - M_{n,k,s}(i-1,j) \\ b = M_{n,k,s}(i,j+1) - M_{n,k,s}(i,j-1) \end{cases} \tag{5}$$

Then, the distribution of the image gradient in 8 directions [49] can be determined based on $\theta_k(i,j)$ and $f_k(i,j)$. After the calculation along eight directions for each subregion, a 128-element array can be obtained, i.e. $d \times d \times 8$ when $d = 4$. Such a 128-element array is then used as the feature description vector ($\overrightarrow{d_f}$) of the feature points [33]. Due to the statistical determination of the array over a wide area, the matching robustness is maintained.

2.4 Feature point matching

The position information (i.e. the pixel coordinates of the feature points) and feature description vector of each feature point in the reference images are denoted as $\overrightarrow{p_{v1}}$ and $\overrightarrow{d_{f1}}$, respectively. The corresponding ones in registered image are denoted as $\overrightarrow{p_{v2}}$ and $\overrightarrow{d_{f2}}$. Then, the difference ($\Delta$) between the feature description vectors is

$$\Delta = \sqrt{\left|\overrightarrow{d_{f1}} - \overrightarrow{d_{f2}}\right|^2}. \tag{6}$$

The smaller the difference ($\Delta$) between the two feature description vectors, the more similar they are. The preset threshold of $\Delta$ is denoted as $\Delta_s$. Clearly, $\Delta_s$ signifies the requirement for similarity between the feature description vectors, thereby determining the number of matching points to be retained. Only when $\Delta < \Delta_s$, the $\overrightarrow{p_{v1}}$, $\overrightarrow{p_{v2}}$, $\overrightarrow{d_{f1}}$ and $\overrightarrow{d_{f2}}$ of the matched pairs are stored for image stitching.

2.5 Image stitching

Based on the matched pairs obtained previously, the imaging stitching process utilizes the Random Sample Consensus (RANSAC) algorithm [50]. RANSAC operates under the fundamental assumption that the sample comprises both accurate data (inliers, data that conform to the model) and abnormal data (outliers, data that deviate from the expected range and do not align with the model), which may, for instance, result from noise [51] and improper measurements, assumptions or calculations. Furthermore, RANSAC also assumes that for a given accurate dataset, the model parameters can be consistently computed.

As shown in Fig. 2(f), a homography matrix ($H$) is employed to depict the perspective transformation of a plane in the real world along with its corresponding image. This matrix is utilized to facilitate the transformation of the image from one viewpoint to another through the perspective transformation process. Therefore, the relationship between the matched pairs can be obtained as follows

$$\begin{bmatrix} x' \\ y' \\ 1 \end{bmatrix} = H \begin{bmatrix} x \\ y \\ 1 \end{bmatrix} \tag{7}$$

where $(x, y)$ represents the pixel coordinates of the matched points in the reference image. $(x', y')$ represents the pixel coordinates of the matching points in the registered image. $H$ can be expressed as

$$H = \begin{bmatrix} \cos\theta & -\sin\theta & t_x \\ \sin\theta & \cos\theta & t_y \\ 0 & 0 & 1 \end{bmatrix} \tag{8}$$

where $\theta$ represents the angular difference between the reference image and the registered image. $t_x$ and $t_y$ represent the translational difference between the reference image and the registered image. All the quantities in Eq. (8) are determined through the matched image pairs.

## 3. Results of stitching on two images

In this section, we present a performance evaluation of the LP-SIFT algorithm, along with comparative results from other algorithms such as SIFT, ORB, BRISK, and SURF. Table 1 presents different hardware and software resources used in this study. The code of the LP-SIFT algorithm has been written in MATLAB 2021a with parallel computing. The code of the SIFT[52] downloaded from the internet. The code of the ORB, SURF, and BRISK are from the Computer Vision Toolbox of the MATLAB.

To ensure consistent comparison across different scenarios, the SIFT, ORB, BRISK, SURF, and LP-SIFT algorithms were utilized to compute feature points and feature description vectors in the two images. Subsequently, the RANSAC algorithm was employed to stitch the images together.

3.1 Datasets

To evaluate the performance of LP-SIFT algorithm, a range of datasets that encompass various scenarios, pixel sizes and different levels of distortions have been studied. These datasets include a rich assortment of illumination conditions and structural features in both natural and artificial environments. In this study, two datasets have been used: Dataset-A and Dataset-B. Dataset-A contains three image pairs commonly adopted in relevant researches, including: (1) mountain [53], (2) street view[54], (3) terrain [55]. Dataset-B dataset is captured by a camera of mobile phone (PGKM10, OnePlus) with a resolution of 6 Mega-Pixels. It contains three image pairs: (1) building, (2) campus view (translation), and (3) campus view (rotation). The image pairs of the datasets are shown in Fig. 3 as examples.

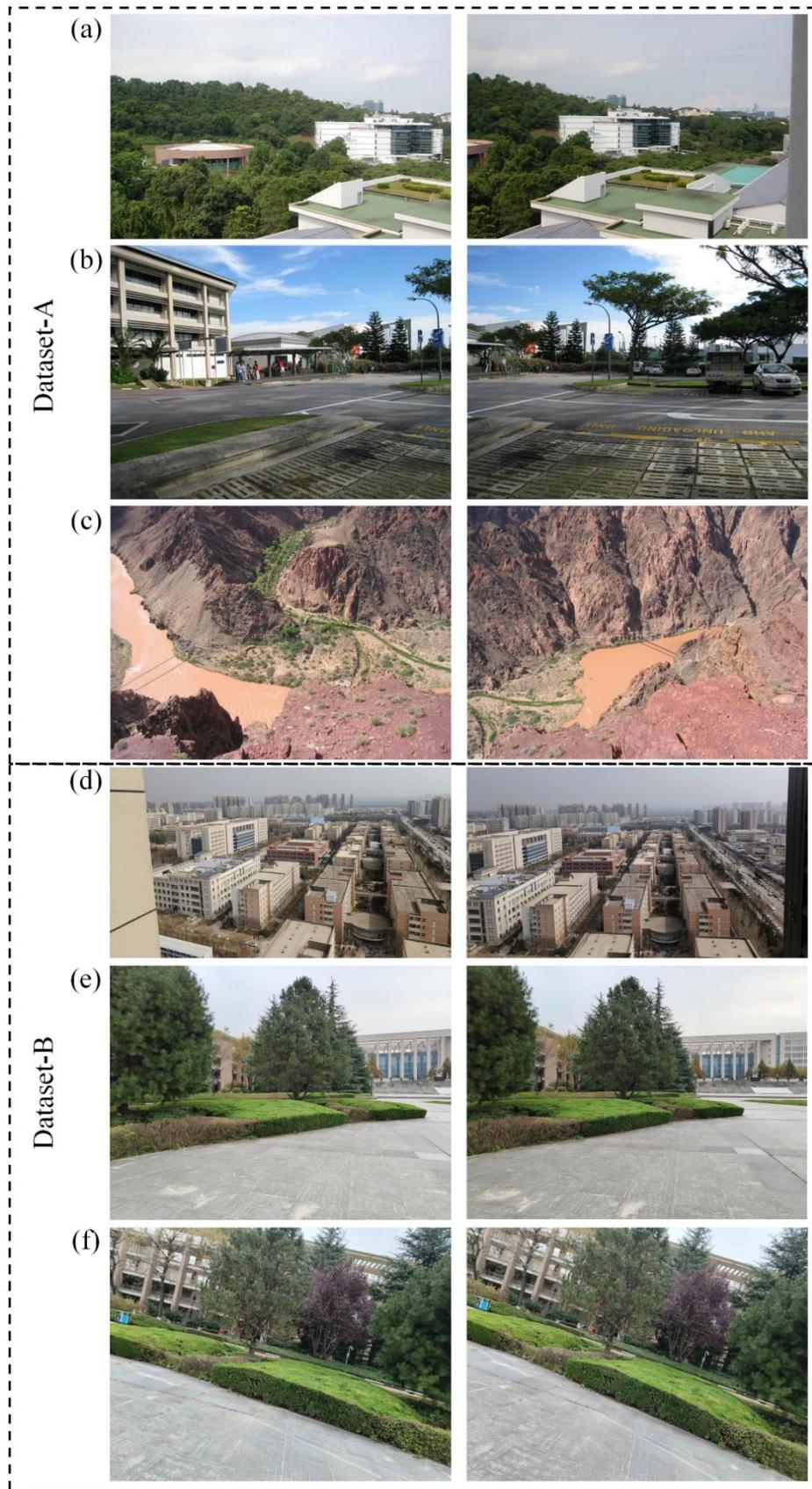

Figure 3. Datasets. Dataset-A: **(a)** mountain [53] dataset image pair, **(b)** street view [54] dataset image pair, **(c)** terrain [55] dataset image pair, Dataset-B: **(d)** building dataset image pair, **(e)** campus view dataset (translation) image pair, (f)campus view dataset (rotation) image pair.

Table 1. Hardware and software specifications

| | | |
|---|---|---|
| Hardware | Operation system | Windows 11 64-bit operating system |
| | Processor | Intel Core i9-12900 |
| | Memory | 64 GB |
| | Graphics card | NVIDIA GeForce RTX 3090 |
| Software | Platform | MATLAB 2021a |
| | Library | Computer Vision Toolbox 10.0 |
| Development environment | SIFT | MATLAB 2021a |
| | ORB | MATLAB 2021a |
| | BRISK | MATLAB 2021a |
| | SURF | MATLAB 2021a |
| | LP-SIFT | MATLAB 2021a |

3.2 Images of small size

Small-sized images are commonly employed in various fields such as medical imaging [18], industrial inspection [56-58], and bridge inspection [59]. Hence, the performance of image stitching, which combines LP-SIFT and RANSAC, is initially assessed for these small-sized images. In our investigation, the images in the datasets of mountain and street view have small sizes. The sizes of the images are 602×400 pixels and 653×490 pixels, respectively. The stitching results are collectively shown in Fig. 4, incorporating those obtained using SIFT, ORB, BRISK, SURF and LP-SIFT. The corresponding parameters of the stitching process are summarized in Table 2. It is evident that all the five feature point detection algorithms, when combined with RANSAC can successfully stitch the two images with comparable quality. However, there are significant differences in computation times. It should be noted that the computation time encompasses feature point detection, feature description vectors calculation, pair matching and image stitching.

For the dataset of mountain, the computation times are 101.21s for SIFT, 0.71s for ORB, 1.30s for BRISK, 0.85s for SURF, and 1.16s for LP-SIFT. ORB takes the least time, while SIFT takes the most time. The comparison can be more visible from Fig. 5. However, for the dataset of street view, the computation times are 226.62s for SIFT, 2.27s for ORB, 2.22s for BRISK, 2.54s for SURF, and 2.05s for LP-SIFT. LP-SIFT takes the least time, while SIFT takes the most time. Furthermore, the stitching outcomes produced by the SIFT algorithm exhibit notable misalignment at the seams, whereas the other algorithms maintain alignment without noticeable discrepancies. In summary, SIFT is the most computationally intensive, with occasional misalignment in stitching results, while stitching time of LP-SIFT, ORB, BRISK, and SURF are on the same level.

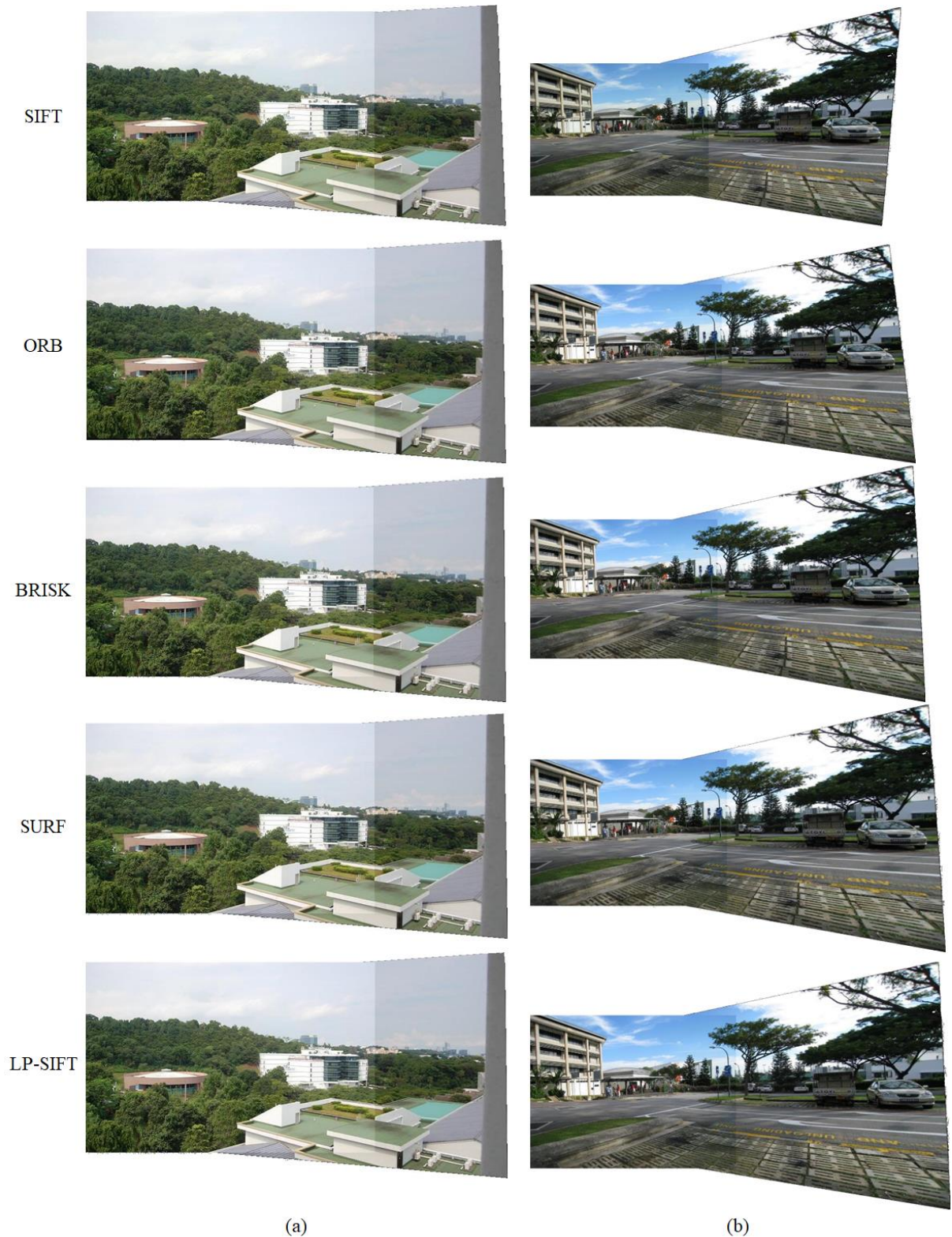

Figure 4. Stitching results of mountain dataset and street dataset. (a) Dataset of mountain stitched by SIFT, ORB, BRISK, SURF, and LP-SIFT respectively. (b) Dataset of street view stitched by SIFT, ORB, BRISK, SURF, and LP-SIFT respectively.

3.3 Images of medium size

For the purpose of high-quality visual presentation of photos [60], videos [61] and other scenes, images of 1080P (1080×1920 pixels) are commonly utilized. In our dataset, the image pairs in the datasets of terrain and building have medium size. The former has a size of 1024×768 pixels and the latter are 1080×1920 pixels.

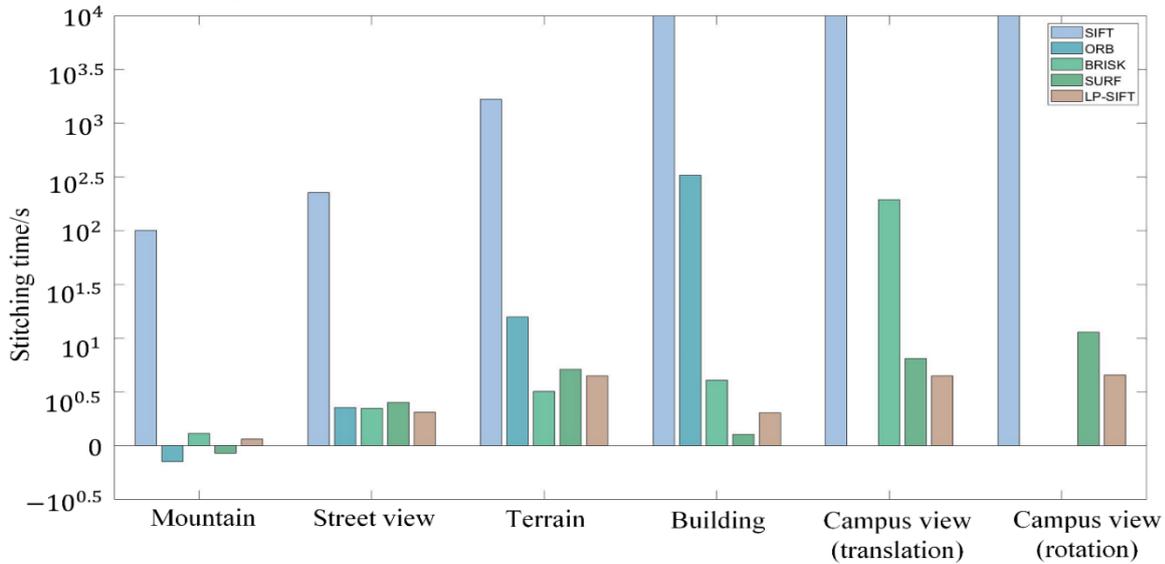

Figure 5. Comparison of the stitching times of 5 algorithms for different datasets

The stitching results are collectively shown in Fig. 6, incorporating those obtained using SIFT, ORB, BRISK, SURF and LP-SIFT. The corresponding parameters of the stitching process are summarized in Table 2. For terrain dataset, the five feature point detection algorithms, when combined with RANSAC, can successfully stitch the two images. The differences are, SIFT algorithm takes up to 1674.87 s to accomplish the stitching. In contrast, ORB, BRISK, SURF and LP-SIFT algorithms take 15.77 s, 3.20s, 5.16 s and 4.47 s respectively.

For the dataset of building, only four feature point detection algorithms, when combined with RANSAC, can successfully stitch the two images. The differences are, ORB algorithm takes 327.25 s, BRISK algorithm takes 4.08s, SURF algorithm takes 1.28 s, and LP-SIFT algorithm takes 2.03 s. Since the SIFT algorithm takes more than $10^4$ s but doesn't return any matching result, the computation is terminated without returning a stitching time. In this computation set, the BRISK, SURF and LP-SIFT algorithms require significantly less time than SIFT and ORB, which experiences a notable increase in the number of feature points as the image size is enlarged.

3.4 Images of large size with translational displacement

Large-sized images are frequently employed in mobile photography [62, 63], satellite remote sensing [64, 65], UAV aerial photography [23, 66] and other fields. In our dataset, the image pairs in the datasets of campus view (translation) have large size. The size of the images is 3072×4096 pixels.

Table 2. Parameter setup of different feature point detection algorithm. To maintain the consistency for comparison, all the of them were stitched by RANSAC algorithm.

| Name | Image size | | Algorithm | Number of feature points | | Number of matched pairs | Window size ($L$) | Stitching time (s) |
|---|---|---|---|---|---|---|---|---|
| Mountain | Small image | 602×400 | SIFT | 1496 | 939 | 15 | × | 101.21 |
| | | | ORB | 10505 | 6926 | 1132 | × | 0.71 |
| | | | BRISK | 1416 | 1122 | 75 | × | 1.30 |
| | | | SURF | 638 | 490 | 216 | × | 0.85 |
| | | | LP-SIFT | 487 | 493 | 31 | [32,40] | 1.16 |
| Street view | Small image | 653×490 | SIFT | 1948 | 2726 | 23 | × | 226.62 |
| | | | ORB | 11363 | 15597 | 541 | × | 2.27 |
| | | | BRISK | 2523 | 4104 | 59 | × | 2.22 |
| | | | SURF | 933 | 1149 | 137 | × | 2.54 |
| | | | LP-SIFT | 811 | 812 | 59 | [32,40] | 2.05 |
| Terrain | Medium image | 1024×768 | SIFT | 9495 | 10368 | 134 | × | 1674.87 |
| | | | ORB | 3182 | 3182 | 2224 | × | 15.77 |
| | | | BRISK | 8149 | 8306 | 95 | × | 3.20 |
| | | | SURF | 2883 | 3037 | 204 | × | 5.16 |
| | | | LP-SIFT | 1847 | 1837 | 29 | [32,64] | 4.47 |
| Building | Medium image | 1080×1920 | SIFT | × | × | × | × | >$10^4$ |
| | | | ORB | 107612 | 108452 | 9720 | × | 327.25 |
| | | | BRISK | 14660 | 15428 | 605 | × | 4.08 |
| | | | SURF | 6123 | 5985 | 1780 | × | 1.28 |
| | | | LP-SIFT | 532 | 484 | 17 | [100,128] | 2.03 |
| Campus view (translation) | Large image | 3072×4096 | SIFT | × | × | × | × | >$10^4$ |
| | | | ORB | 1025750 | 927050 | Over size | × | × |
| | | | BRISK | 104981 | 94657 | 3299 | × | 195.44 |
| | | | SURF | 27790 | 25465 | 6056 | × | 6.52 |
| | | | LP-SIFT | 418 | 403 | 29 | [256,512] | 4.49 |
| Campus view (rotation) | Large image | 3072×4096 | SIFT | × | × | × | × | >$10^4$ |
| | | | ORB | 1326389 | 1332929 | Over size | × | × |
| | | | BRISK | 158247 | 164035 | Over size | × | × |
| | | | SURF | 47568 | 47678 | 11293 | × | 11.42 |
| | | | LP-SIFT | 422 | 429 | 22 | [256,512] | 4.58 |

The stitching results are collectively shown in Fig. 7(a), incorporating those obtained using BRISK, SURF and LP-SIFT. The corresponding parameters of the stitching process are summarized in Table 2. For campus view (translation) datasets, only three feature point detection algorithms, when combined with RANSAC, can successfully stitch the two images.

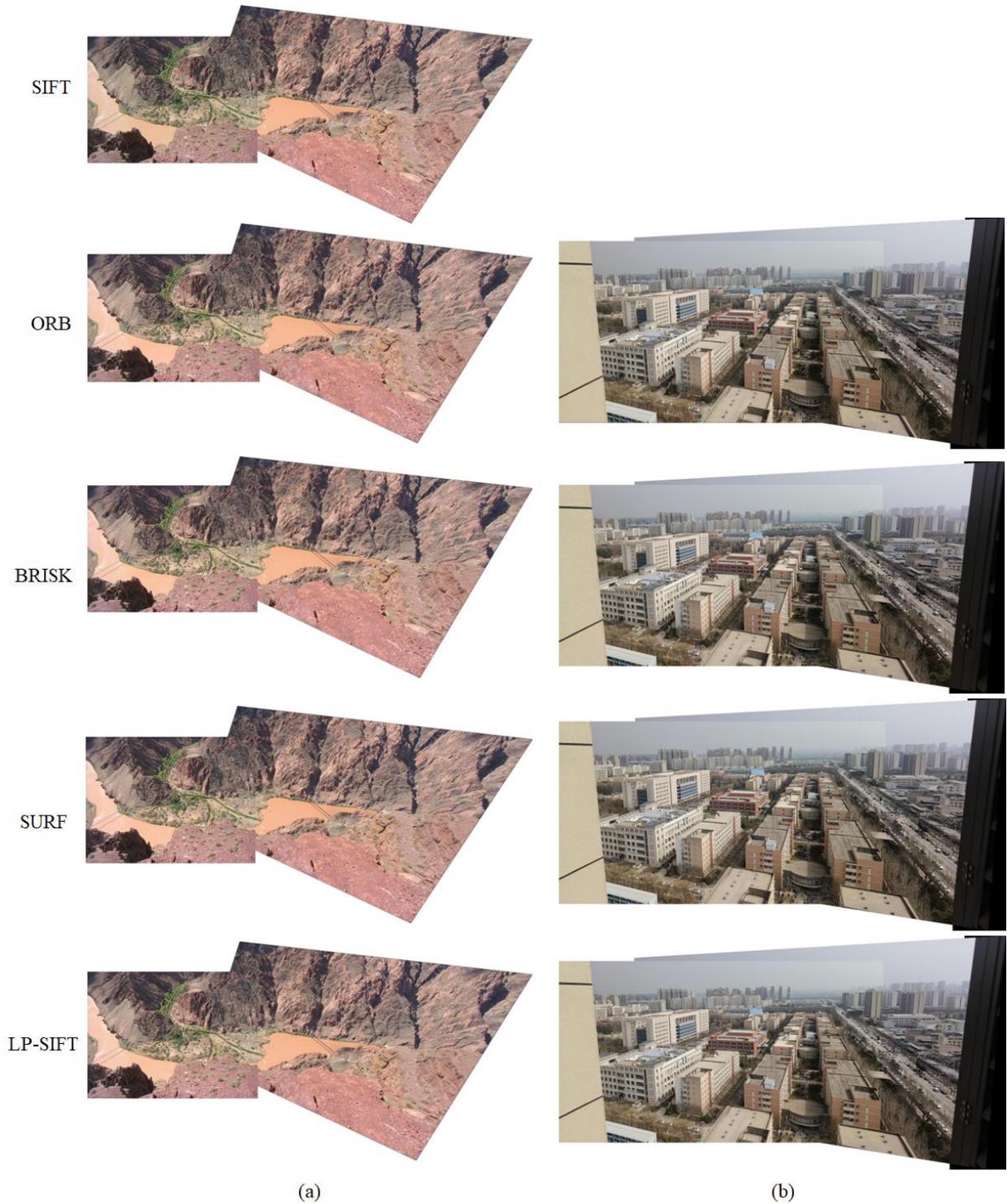

Figure 6. Stitching results of terrain dataset and building dataset. (a) Dataset of terrain stitched by SIFT, ORB, BRISK, SURF, and LP-SIFT respectively. (b) Dataset of building stitched by ORB, BRISK, SURF, and LP-SIFT respectively.

The differences are, BRISK algorithm takes 195.44s, SURF algorithm takes 6.52 s, and LP-SIFT algorithm takes 4.49s. Since the execution time for SIFT exceeded $10^4$ s without yielding any results, the computation was halted at that point. On the other hand, the ORB algorithm detected an excessive number of feature points, surpassing the computer's memory capacity and leading to a stitching failure. Among the three algorithms, LP-SIFT exhibits orders of improvement on stitching efficiency. One key advantage of LP-SIFT is its adjustable interrogation window size, allowing for the control of the number of feature points even in extremely large images with high signal-to-noise ratio (SNR). By applying a larger interrogation window size, LP-SIFT can effectively limit the number of feature points, resulting in a faster stitching process compared to other algorithms. This enhanced efficiency makes LP-SIFT a compelling choice for image stitching tasks, especially in scenarios where computational resources are limited.

3.5 Images of large size with rotational displacement

SIFT algorithm has rotation invariance. In this section, we hope to demonstrate LP-SIFT algorithm also reserve rotation invariance feature. The image pairs in the datasets of campus view (rotation) with a large size of 3072×4096 pixels are applied. The image pairs are captured separately by the same camera, not artificially rotated from each other. SIFT, ORB, BRISK, SURF and LP-SIFT algorithms were employed to detect feature points and feature description vectors. RANSAC algorithm is further used to stitch the images. Fig. 7(b) depicts the stitching results achieved by combining the SURF and LP-SIFT feature point detection algorithms with RANSAC, along with the stitching parameters summarized in Table 2. It is evident that both feature point detection algorithms can successfully stitch the two images with comparable quality. However, SURF algorithm takes 11.42 s, while LP-SIFT algorithm takes 4.58 s, which is only 40% of the computation time by SURF. Similar as in section 3.4, SIFT algorithm requires an unacceptable long time for stitching, while ORB and BRISK algorithm fail in stitching due to the overflow of memory. Therefore, LP-SIFT shows the capability of stitching images with rotational displacements, particularly fast for the images with large size.

3.6 Discussion

The test results demonstrate that compared to the original SIFT method, the speed of feature point detection can be improved by 109 times for small size images and by orders of magnitude for larger images. The time consumption in SIFT is primarily attributed to the calculation of feature description vectors, particularly when dealing with many detected feature points, especially in the case of large images.

Although ORB and BRISK show good robustness in different scenarios, they face challenges due to the rapid increase in feature points with image size. This may not be a big deal if the computational resource is sufficient large, e.g. a commercial workstation. But it could be an obstacle for the applications in a portable computation system. SURF algorithm is applicable for stitching images of different sizes, but the overall efficiency is relatively lower than LP-SIFT.

Relative to SIFT, ORB, BRISK, and SURF, it is flexible for LP-SIFT to adjust the interrogation window size to determine the multi-scale local peaks. Therefore, the number of feature points can be well controlled without a significant increase with the image size, which is then advantageous to replace the Gaussian pyramid and difference pyramid feature point detection.

One may need to note, the algorithm of LP-SIFT is programed in Matlab without acceleration by GPU. If LP-SIFT is developed by C/C++ with GPU acceleration, the computation efficiency can be significantly promoted.

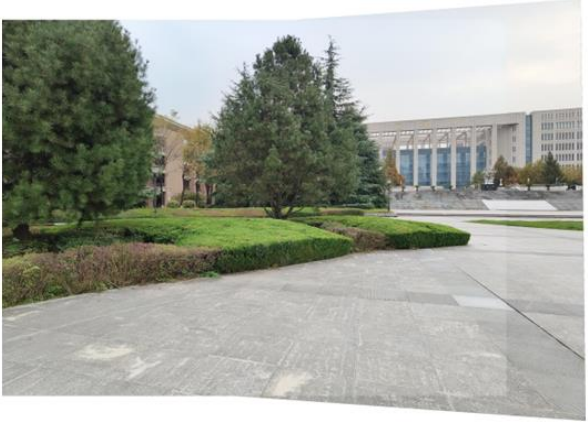

BRISK

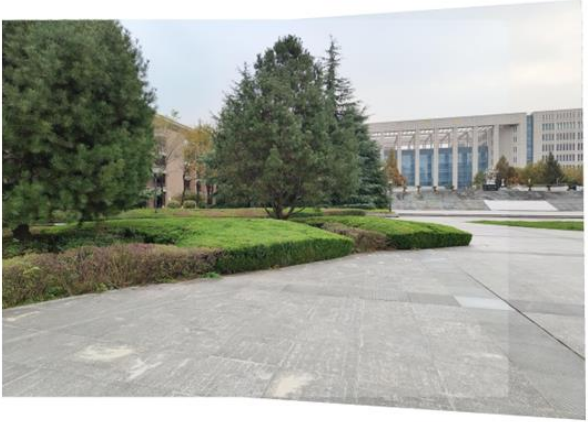
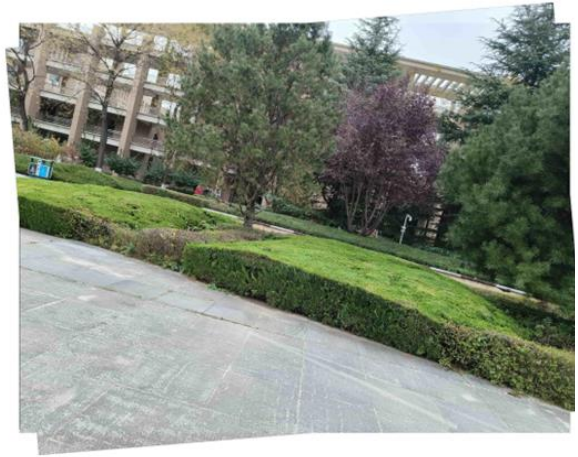

SURF

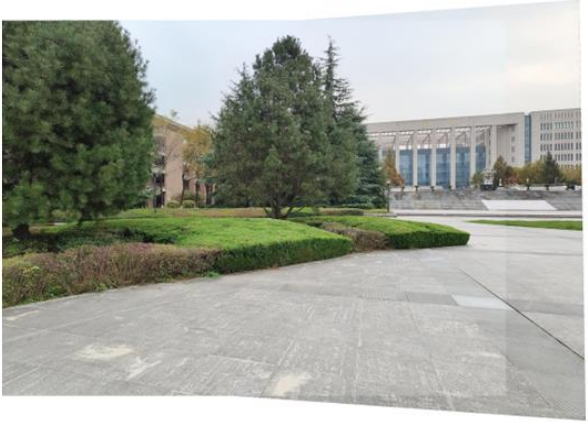
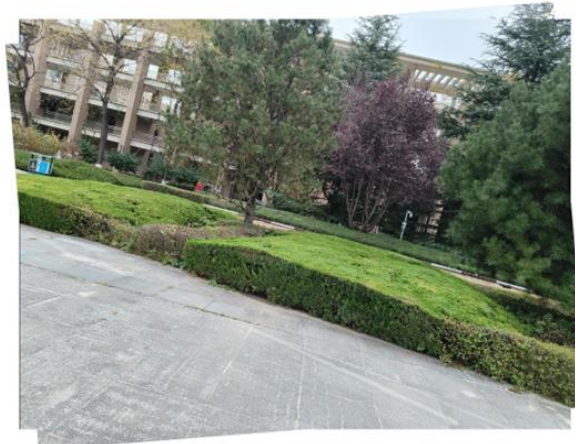

LP-SIFT

(a) (b)

Figure 7. Stitching results of campus view dataset. (a) Dataset of campus view (translation) stitched by BRISK, SURF, and LP-SIFT respectively. (b) Dataset of campus view (rotation) stitched by SURF, and LP-SIFT respectively.

## 4. Mosaic of multiple images without prior knowledge

In various application scenes, e.g. criminal investigation [67], remote sensing monitoring [65] and UAV aerial photography [66] etc., images are very probably fragmented with unknown positions, angles and sequences that need to be restored. Thus, the mosaic strategy of stitching multiple images, which is more complex than the two-image case, is also crucial. Here we propose a mosaic strategy for combining multiple images without prior knowledge.

As illustrated in Fig. 8, the first step involves using LP-SIFT to compute the homography matrix between each pair of images within a given dataset, which is stored to the matrix ($H_M$). This aims to find the transformation relationship between individual image pairs. The process can be parallelized using CPU computing to save computation time.

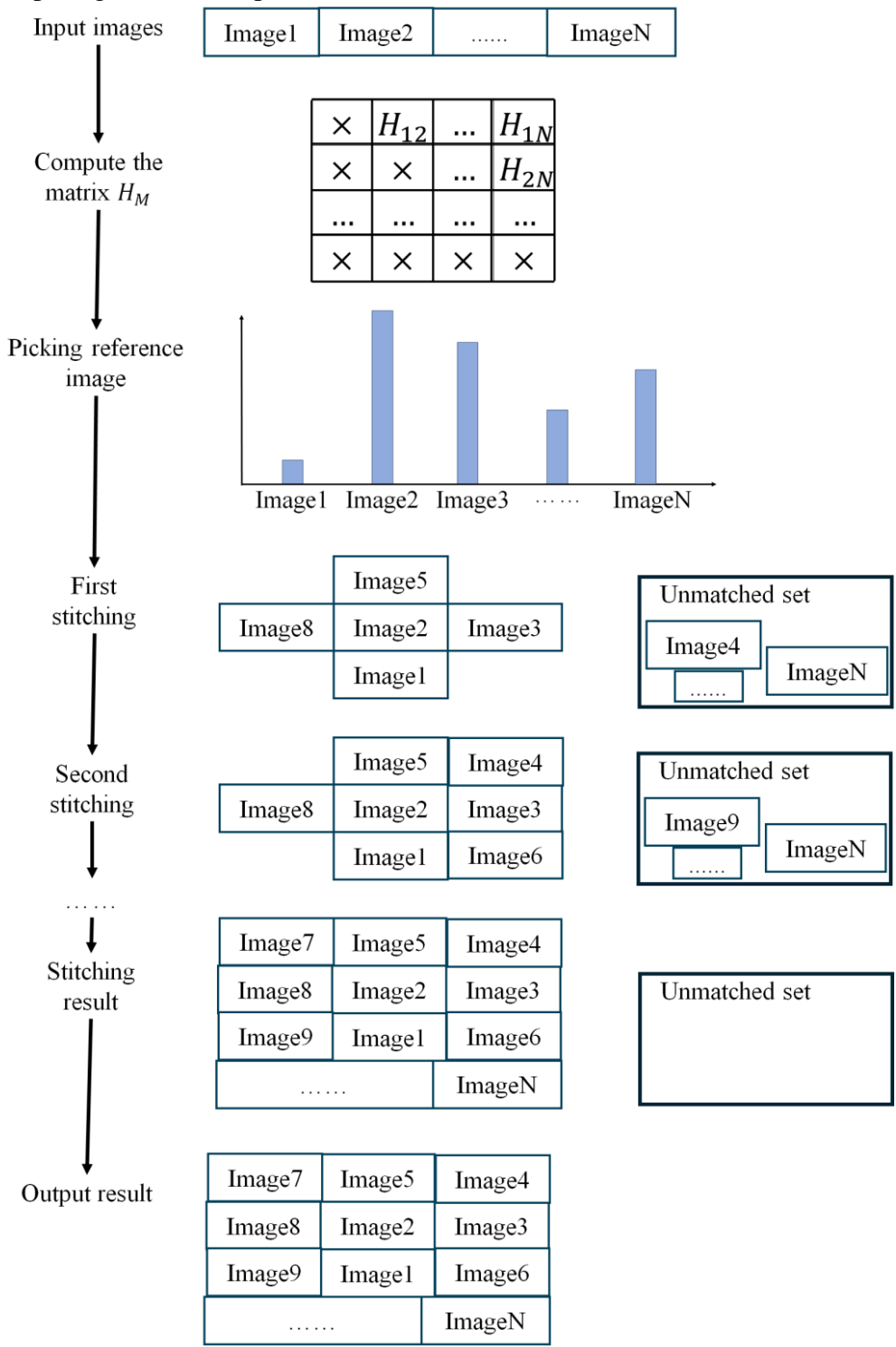

Figure 8. Schematic diagram of LP-SIFT image mosaic of multiple images without prior knowledge

Subsequently, the number of nonzero elements in each row or column of $H_M$ is counted, and the reference image is selected based on the row or column with the highest count of nonzero elements. This process guarantees the reference image with the most neighbor images (the images can be stitched with the reference image) can be stitched first to reduce the subsequent iterations and save Mosaic time. The images matched with the reference image are stitched in this round, leaving the unmatched images stored in the unmatched set. Then, in the second round, another reference image is selected from the unmatched set according to $H_M$. By repeating the process above, the Mosaic is finally terminated until the number in the unmatched set reaches 0.

Fig. 9(a) depicts the original image captured in this experiment. The image size is 6400×4270. Fig. 9(b) shows the fragmented images randomly divided from the original image. The approach outlined in Fig. 8 is employed to seamlessly merge the image fragments shown in Fig. 9(b). As shown in Fig. 9(c), the result is highly coincident to the original image in Fig. 9(a). The stitching time is 158.94 s, which is acceptable for a wide range of applications.

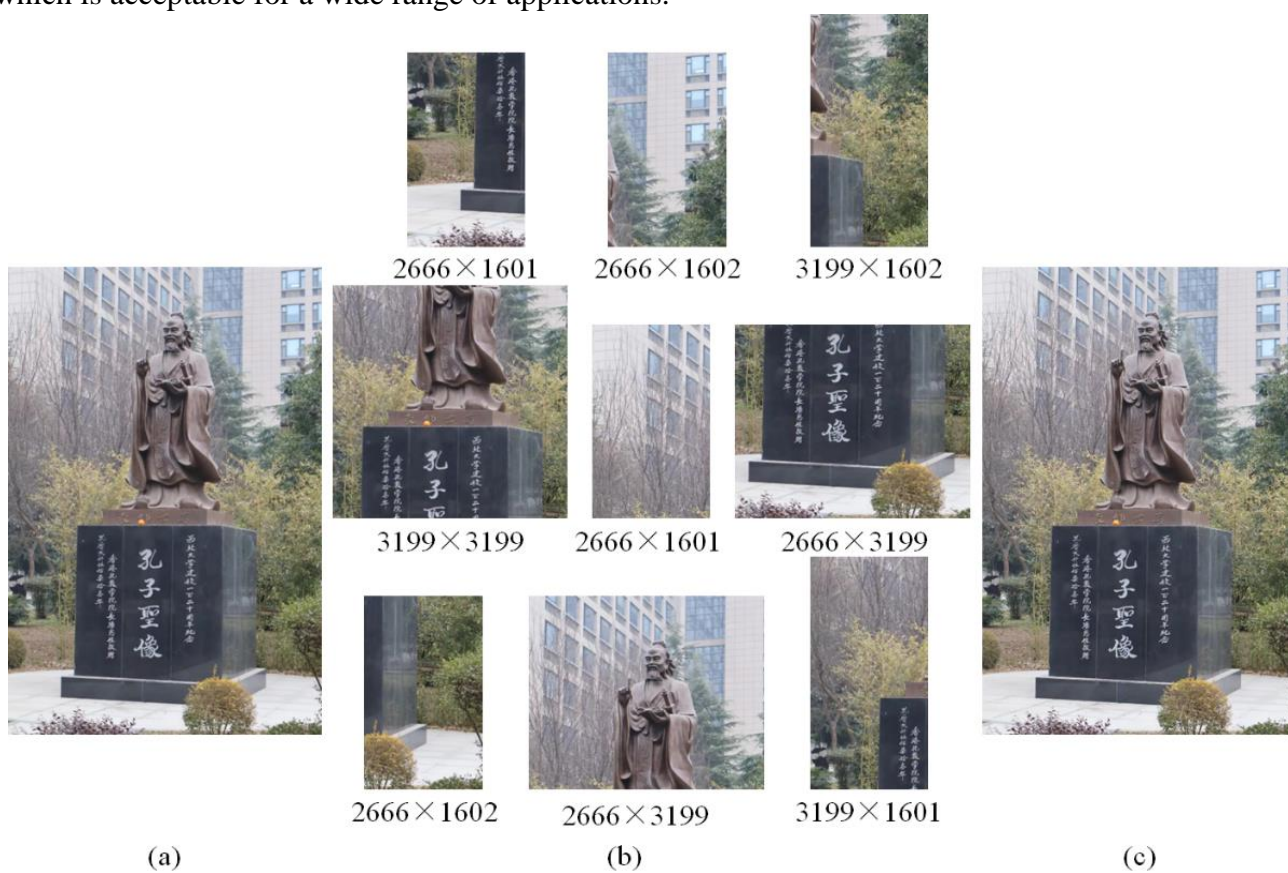

Figure 9. Mosaic of multiple images without prior knowledge. (a) Original image, the image size is 6400×4270. (b) The original image is stitched into different sizes and its position is shuffled, and its size is marked below the image. (c) stitching result, and the stitching time is 158.94 s.

## 5. Conclusions

In this study, we propose a fast feature point detection algorithm, namely Local-peak scale-invariant feature transform (LP-SIFT), which integrates the concept of local extremal points or image peaks at the multiscale level with SIFT. By integrating LP-SIFT and RANSAC in image stitching, significant improvements in stitching speed are achieved compared to the original SIFT method. Furthermore, LP-SIFT evaluated against ORB, BRISK, and SURF for stitching images of varying sizes. Due to its adaptability in adjusting the interrogation window size, LP-SIFT demonstrates minimal increase in the number of detected feature points with increasing image size, resulting in noticeable reduction in stitching time particularly for large-scale cases. Additionally, we also provide a strategy for seamlessly stitching multiple images using LP-SIFT without prior knowledge. It is anticipated that LP-SIFT will contribute to diverse application scenarios such as terrain mapping, biological analysis, and even criminal investigations.

**Author Contributions**
**Hao Li:** Investigation, Visualization, Writing- Original draft preparation, Software. **Lipo Wang:** Methodology, Writing- Reviewing and Editing. **Tianyun Zhao:** Writing- Reviewing and Editing. **Wei Zhao:** Supervision, Conceptualization, Methodology, Data curation, Validation, Writing- Reviewing and Editing.
**Funding:** National Natural Science Foundation of China (51927804, 61378083, 61775181).
**Conflicts of Interest:** The authors declare no conflict of interest.
**Statement:** During the preparation of this work the author(s) used ChatGPT 4.0 in order to check essay presentation, grammar, and spelling. After using this tool, the authors reviewed and edited the content as needed and take full responsibility for the content of the publication.

**References**
[1] P. Deshmukh and P. Paikrao. A Review of Various Image Mosaicing Techniques, 2019 Innovations in Power and Advanced Computing Technologies (I-Pact), 2019. https://doi.org/10.1109/i-pact44901.2019.8960220.
[2] W. T. Lee, H. I. Chen, M. S. Chen, I. C. Shen and B. Y. Chen. High-Resolution 360 Video Foveated Stitching for Real-Time Vr, Computer Graphics Forum, 2017: 115-123. https://doi.org/10.1111/cgf.13277.
[3] S. Saeed, M. U. Kakli, Y. Cho, J. Seo and U. Park. A High-Quality Vr Calibration and Real-Time Stitching Framework Using Preprocessed Features, IEEE Access, 2020: 190300-190311. https://doi.org/10.1109/access.2020.3031413.
[4] P. Dang, J. Zhu, Y. Zhou, Y. Rao, J. You, J. Wu, M. Zhang and W. Li. A 3d-Panoramic Fusion Flood Enhanced Visualization Method for Vr, Environmental Modelling & Software, 2023. https://doi.org/10.1016/j.envsoft.2023.105810.
[5] Z. Liu and S. Chang. A Study of Digital Exhibition Visual Design Led by Digital Twin and Vr Technology, Measurement: Sensors, 2024. https://doi.org/10.1016/j.measen.2023.100970.
[6] T. Greibe, T. A. Anhøj, L. S. Johansen and A. Han. Quality Control of Jeol Jbx-9500fsz E-Beam Lithography System in a Multi-User Laboratory, Microelectronic Engineering, 2016: 25-28. https://doi.org/10.1016/j.mee.2016.02.003.
[7] J. Pan, W. Liu, P. Ge, F. Li, W. Shi, L. Jia and H. Qin. Real-Time Segmentation and Tracking of Excised Corneal Contour by Deep Neural Networks for Dalk Surgical Navigation, Computer Methods and Programs in Biomedicine, 2020. https://doi.org/10.1016/j.cmpb.2020.105679.

[8] Y. Wang and J. Yang. Origin of Organic Matter Pore Heterogeneity in Oil Mature Triassic Chang-7 Mudstones, Ordos Basin, China, International Journal of Coal Geology, 2024. https://doi.org/10.1016/j.coal.2024.104458.

[9] Q. Zhou and Z. Zhou. Web-Based Mixed Reality Video Fusion with Remote Rendering, Virtual Reality & Intelligent Hardware, 2023: 188-199. https://doi.org/10.1016/j.vrih.2022.03.005.

[10] Z. He, Z. He, S. Li, Y. Yu and K. Liu. A Ship Navigation Risk Online Prediction Model Based on Informer Network Using Multi-Source Data, Ocean Engineering, 2024. https://doi.org/10.1016/j.oceaneng.2024.117007.

[11] B. Wang, S. Gou, K. Di, W. Wan, M. Peng, C. Zhao, Y. Zhang and B. Xie. Rock Size-Frequency Distribution Analysis at the Zhurong Landing Site Based on Navigation and Terrain Camera Images Along the Entire Traverse, Icarus, 2024. https://doi.org/10.1016/j.icarus.2024.116001.

[12] M. Cao, L. Zheng, W. Jia and X. Liu. Constructing Big Panorama from Video Sequence Based on Deep Local Feature, Image and Vision Computing, 2020. https://doi.org/10.1016/j.imavis.2020.103972.

[13] W. Lyu, Z. Zhou, L. Chen and Y. Zhou. A Survey on Image and Video Stitching, Virtual Reality & Intelligent Hardware, 2019: 55-83. https://doi.org/10.3724/sp.J.2096-5796.2018.0008.

[14] Q. Wang, F. Reimeier and K. Wolter. Efficient Image Stitching through Mobile Offloading, Electronic Notes in Theoretical Computer Science, 2016: 125-146. https://doi.org/10.1016/j.entcs.2016.09.027.

[15] R. Torres, G. Mahalingam, D. Kapner, E. T. Trautman, T. Fliss, S. Seshamani, E. Perlman, R. Young, S. Kinn, J. Buchanan, M. M. Takeno, W. Yin, D. J. Bumbarger, R. P. Gwinn, J. Nyhus, E. Lein, S. J. Smith, R. C. Reid, K. A. Khairy, S. Saalfeld, F. Collman and N. Macarico da Costa. A Scalable and Modular Automated Pipeline for Stitching of Large Electron Microscopy Datasets, eLife, 2022. https://doi.org/10.7554/eLife.76534.

[16] B. Ma, T. Zimmermann, M. Rohde, S. Winkelbach, F. He, W. Lindenmaier and K. E. J. Dittmar. Use of Autostitch for Automatic Stitching of Microscope Images, Micron, 2007: 492-499. https://doi.org/10.1016/j.micron.2006.07.027.

[17] F. Yang, Z.-S. Deng and Q.-H. Fan. A Method for Fast Automated Microscope Image Stitching, Micron, 2013: 17-25. https://doi.org/10.1016/j.micron.2013.01.006.

[18] F. Yang, Y. He, Z. S. Deng and A. Yan. Improvement of Automated Image Stitching System for Dr X-Ray Images, Computers in Biology and Medicine, 2016: 108-114. https://doi.org/10.1016/j.compbiomed.2016.01.026.

[19] J.-H. Seo, S. Yang, M.-S. Kang, N.-G. Her, D.-H. Nam, J.-H. Choi and M. H. Kim. Automated Stitching of Microscope Images of Fluorescence in Cells with Minimal Overlap, Micron, 2019. https://doi.org/10.1016/j.micron.2019.102718.

[20] Z. Lei, X. Liu, L. Zhao, L. Chen, Q. Li, T. Yuan and W. Lu. A Novel 3d Stitching Method for Wli Based Large Range Surface Topography Measurement, Optics Communications, 2016: 435-447. https://doi.org/10.1016/j.optcom.2015.09.074.

[21] P. Yang, S.-w. Ye and Y.-f. Peng. Three-Dimensional Profile Stitching Measurement for Large Aspheric Surface During Grinding Process with Sub-Micron Accuracy, Precision Engineering, 2017: 62-71. https://doi.org/10.1016/j.precisioneng.2016.07.005.

[22]W. Y. Kim, B. W. Seo, S. H. Lee, T. G. Lee, S. Kwon, W. S. Chang, S.-H. Nam, N. X. Fang, S. Kim and Y. T. Cho. Quasi-Seamless Stitching for Large-Area Micropatterned Surfaces Enabled by Fourier Spectral Analysis of Moiré Patterns, Nature Communications, 2023. https://doi.org/10.1038/s41467-023-37828-8.

[23]A. Feng, C. N. Vong, J. Zhou, L. S. Conway, J. Zhou, E. D. Vories, K. A. Sudduth and N. R. Kitchen. Developing an Image Processing Pipeline to Improve the Position Accuracy of Single Uav Images, Computers and Electronics in Agriculture, 2023. https://doi.org/10.1016/j.compag.2023.107650.

[24]S. Feng, M. Gao, X. Jin, T. Zhao and F. Yang. Fine-Grained Damage Detection of Cement Concrete Pavement Based on Uav Remote Sensing Image Segmentation and Stitching, Measurement, 2024. https://doi.org/10.1016/j.measurement.2023.113844.

[25]X. Wang, N. He, C. Hong, Q. Wang and M. Chen. Improved Yolox-X Based Uav Aerial Photography Object Detection Algorithm, Image and Vision Computing, 2023. https://doi.org/10.1016/j.imavis.2023.104697.

[26]W. Zeng, Q. Deng, X. Zhao, D. Li and X. Min. A Method for Stitching Remote Sensing Images with Delaunay Triangle Feature Constraints, Geocarto International, 2023. https://doi.org/10.1080/10106049.2023.2285356.

[27]T. Rui, Y. Hu, C. Yang, D. Wang and X. Liu. Research on Fast Natural Aerial Image Mosaic, Computers & Electrical Engineering, 2021. https://doi.org/10.1016/j.compeleceng.2021.107007.

[28]D. Ghosh and N. Kaabouch. A Survey on Image Mosaicing Techniques, Journal of Visual Communication and Image Representation, 2016: 1-11. https://doi.org/10.1016/j.jvcir.2015.10.014.

[29]Z. Ma and S. Liu. A Review of 3d Reconstruction Techniques in Civil Engineering and Their Applications, Advanced Engineering Informatics, 2018: 163-174. https://doi.org/10.1016/j.aei.2018.05.005.

[30]P. Deshmukh, P. J. I. i. P. Paikrao and A. C. Technologies. A Review of Various Image Mosaicing Techniques, 2019: 1-4.

[31]M. Z. Bonny and M. S. Uddin. Feature-Based Image Stitching Algorithms, 2016 International workshop on computational intelligence (IWCI), 2016: 198-203.

[32]C. Harris and M. Stephens. A Combined Corner and Edge Detector, Procedings of the Alvey Vision Conference 1988, 1988: 23.21-23.26. https://doi.org/10.5244/c.2.23.

[33]D. G. Lowe. Distinctive Image Features from Scale-Invariant Keypoints, International Journal of Computer Vision, 2004: 91-110. https://doi.org/10.1023/b:Visi.0000029664.99615.94.

[34]E. Rosten and T. Drummond. Machine Learning for High-Speed Corner Detection, Computer Vision–ECCV 2006: 9th European Conference on Computer Vision, Graz, Austria, May 7-13, 2006. Proceedings, Part I 9, 2006: 430-443.

[35]H. Bay, A. Ess, T. Tuytelaars and L. Van Gool. Speeded-up Robust Features (Surf), Computer Vision and Image Understanding, 2008: 346-359. https://doi.org/10.1016/j.cviu.2007.09.014.

[36]E. Rublee, V. Rabaud, K. Konolige and G. Bradski. Orb: An Efficient Alternative to Sift or Surf, 2011 International conference on computer vision, 2011: 2564-2571.

[37]S. Leutenegger, M. Chli and R. Y. Siegwart. Brisk: Binary Robust Invariant Scalable Keypoints, 2011 International Conference on Computer Vision, 2011: 2548-2555. https://doi.org/10.1109/iccv.2011.6126542.

[38] P. F. Alcantarilla, A. Bartoli and A. J. Davison. Kaze Features, Computer Vision–ECCV 2012: 12th European Conference on Computer Vision, Florence, Italy, October 7-13, 2012, Proceedings, Part VI 12, 2012: 214-227.

[39] J. N. Pablo F. Alcantarilla, Adrien Bartoli. Fast Explicit Diffusion for Accelerated Features in Nonlinear Scale Spaces, British Machine Vision Conference, 2013.

[40] G. Li, T. Li, F. Li and C. Zhang. Nervestitcher: Corneal Confocal Microscope Images Stitching with Neural Networks, Computers in Biology and Medicine, 2022. https://doi.org/10.1016/j.compbiomed.2022.106303.

[41] F. Zhu, J. Li, B. Zhu, H. Li and G. Liu. Uav Remote Sensing Image Stitching Via Improved Vgg16 Siamese Feature Extraction Network, Expert Systems with Applications, 2023. https://doi.org/10.1016/j.eswa.2023.120525.

[42] N. ul-Huda, H. Ahmad, A. Banjar, A. O. Alzahrani, I. Ahmad and M. S. Naeem. Image Synthesis of Apparel Stitching Defects Using Deep Convolutional Generative Adversarial Networks, Heliyon, 2024. https://doi.org/10.1016/j.heliyon.2024.e26466.

[43] Z. Wu and H. Wu. Improved Sift Image Feature Matching Algorithm, 2022 2nd International Conference on Computer Graphics, Image and Virtualization (ICCGIV), 2022: 223-226. https://doi.org/10.1109/iccgiv57403.2022.00051.

[44] W. Gan, Z. Wu, M. Wang and X. Cui. Image Stitching Based on Optimized Sift Algorithm, 2023 5th International Conference on Intelligent Control, Measurement and Signal Processing (ICMSP), 2023: 1099-1102. https://doi.org/10.1109/icmsp58539.2023.10170989.

[45] X. Li and S. Li. Image Registration Algorithm Based on Improved Sift, 2023 4th International Conference on Electronic Communication and Artificial Intelligence (ICECAI), 2023: 264-267. https://doi.org/10.1109/icecai58670.2023.10176776.

[46] L. Wang and N. Peters. The Length-Scale Distribution Function of the Distance between Extremal Points in Passive Scalar Turbulence, Journal of Fluid Mechanics, 2006. https://doi.org/10.1017/s0022112006009128.

[47] N. Peters and L. Wang. Dissipation Element Analysis of Scalar Fields in Turbulence, Comptes Rendus Mécanique, 2006: 493-506. https://doi.org/10.1016/j.crme.2006.07.006.

[48] L. P. Wang and Y. X. Huang. Multi-Level Segment Analysis: Definition and Application in Turbulent Systems, Journal of Statistical Mechanics: Theory and Experiment, 2015. https://doi.org/10.1088/1742-5468/2015/06/p06018.

[49] D. G. Lowe. Object Recognition from Local Scale-Invariant Features, Proceedings of the seventh IEEE international conference on computer vision, 1999: 1150-1157.

[50] Y. Zhang and Y. Xie. Adaptive Clustering Feature Matching Algorithm Based on Sift and Ransac, 2021 2nd International Conference on Electronics, Communications and Information Technology (CECIT), 2021: 174-179.

[51] M. A. Fischler and R. C. J. C. o. t. A. Bolles. Random Sample Consensus: A Paradigm for Model Fitting with Applications to Image Analysis and Automated Cartography, 1981: 381-395.

[52] H. Wu. Image Stitching. https://github.com/haoningwu3639/ImageStitching; 2021 [accessed Jun 22, 2021].

[53] J. Zaragoza, T.-J. Chin, M. S. Brown and D. Suter. As-Projective-as-Possible Image Stitching with Moving Dlt, Proceedings of the IEEE conference on computer vision and pattern recognition, 2013: 2339-2346.

[54] J. C. Hernandez Zaragoza. As-Projective-as-Possible Image Stitching with Moving Dlt, 2014.

[55] A. V. a. B. Fulkerson. Vlfeat: An Open and Portable Library of Computer Vision Algorithms., 2008. Retrieved from http://www.vlfeat.org/.

[56] D. Zhang, W. Jackson, G. Dobie, G. West and C. MacLeod. Structure-from-Motion Based Image Unwrapping and Stitching for Small Bore Pipe Inspections, Computers in Industry, 2022. https://doi.org/10.1016/j.compind.2022.103664.

[57] S. Chatterjee and K. K. Issac. Viewpoint Planning and 3d Image Stitching Algorithms for Inspection of Panels, NDT & E International, 2023. https://doi.org/10.1016/j.ndteint.2023.102837.

[58] S. Popovych, T. Macrina, N. Kemnitz, M. Castro, B. Nehoran, Z. Jia, J. A. Bae, E. Mitchell, S. Mu, E. T. Trautman, S. Saalfeld, K. Li and H. S. Seung. Petascale Pipeline for Precise Alignment of Images from Serial Section Electron Microscopy, Nature Communications, 2024. https://doi.org/10.1038/s41467-023-44354-0.

[59] R. Xie, J. Yao, K. Liu, X. Lu, Y. Liu, M. Xia and Q. Zeng. Automatic Multi-Image Stitching for Concrete Bridge Inspection by Combining Point and Line Features, Automation in Construction, 2018: 265-280. https://doi.org/10.1016/j.autcon.2018.02.021.

[60] W. Zhu, L. Liu, G. Jiang, S. Yin, S. J. I. T. o. C. Wei and S. f. V. Technology. A 135-Frames/S 1080p 87.5-Mw Binary-Descriptor-Based Image Feature Extraction Accelerator, 2015: 1532-1543.

[61] X. Zhang, H. Sun, S. Chen, N. J. I. T. o. C. Zheng and S. f. V. Technology. Vlsi Architecture Exploration of Guided Image Filtering for 1080p@ 60hz Video Processing, 2016: 230-241.

[62] M. Bordallo-Lopez, O. Silvén, M. Tico and M. Vehviläinen. Creating Panoramas on Mobile Phones, Computational imaging V, 2007: 54-63.

[63] Y. Xiong and K. J. I. T. o. C. E. Pulli. Fast Panorama Stitching for High-Quality Panoramic Images on Mobile Phones, 2010: 298-306.

[64] L. Wang, Y. Zhang, T. Wang, Y. Zhang, Z. Zhang, Y. Yu and L. J. R. S. Li. Stitching and Geometric Modeling Approach Based on Multi-Slice Satellite Images, 2021: 4663.

[65] B. Huang, L. M. Collins, K. Bradbury and J. M. Malof. Deep Convolutional Segmentation of Remote Sensing Imagery: A Simple and Efficient Alternative to Stitching Output Labels, IGARSS 2018-2018 IEEE International Geoscience and Remote Sensing Symposium, 2018: 6899-6902.

[66] M. Ren, J. Li, L. Song, H. Li and T. Xu. Mlp-Based Efficient Stitching Method for Uav Images, IEEE Geoscience and Remote Sensing Letters, 2022: 1-5. https://doi.org/10.1109/lgrs.2022.3141890.

[67] G. Sansoni, M. Trebeschi and F. J. S. Docchio. State-of-the-Art and Applications of 3d Imaging Sensors in Industry, Cultural Heritage, Medicine, and Criminal Investigation, 2009: 568-601.